\documentclass[11pt,a4paper]{article}
\usepackage[hyperref]{acl2019}
\usepackage{times}
\usepackage{latexsym}

\usepackage{graphicx}
\usepackage[export]{adjustbox}
\usepackage{url}
\usepackage{bm}
\usepackage{here}
\usepackage[fleqn]{amsmath}
\usepackage{amssymb}
\usepackage{amsfonts}
\usepackage{multirow}
\usepackage{arydshln}
\usepackage{enumitem}

\usepackage{algorithm}
\usepackage{algpseudocode}
\usepackage{soul}
\usepackage{float}

\usepackage{CJKutf8}

\aclfinalcopy 

\title{A High-Quality Multilingual Dataset for\\Structured Documentation Translation}

\author{
{\bf Kazuma Hashimoto~~~~
Raffaella Buschiazzo~~~~
James Bradbury\thanks{~~Now at Google Brain.}} \\
{\bf Teresa Marshall~~~~
Richard Socher~~~~
Caiming Xiong} \\
Salesforce \\
{\tt \{k.hashimoto,rbuschiazzo,james.bradbury,}\\
{\tt teresa.marshall,rsocher,cxiong\}@salesforce.com}
}

\date{}

\begin{document}
\maketitle

\begin{abstract}
This paper presents a high-quality multilingual dataset for the documentation domain to advance research on localization of structured text.
Unlike widely-used datasets for translation of plain text, we collect XML-structured parallel text segments from the online documentation for an enterprise software platform.
These Web pages have been professionally translated from English into 16 languages and maintained by domain experts, and around 100,000 text segments are available for each language pair.\footnote{Our new dataset is available at \url{https://github.com/salesforce/localization-xml-mt}.}
We build and evaluate translation models for seven target languages from English, with several different copy mechanisms and an XML-constrained beam search.
We also experiment with a non-English pair to show that our dataset has the potential to explicitly enable $17 \times 16$ translation settings.
Our experiments show that learning to translate with the XML tags improves translation accuracy, and the beam search accurately generates XML structures.
We also discuss trade-offs of using the copy mechanisms by focusing on translation of numerical words and named entities.
We further provide a detailed human analysis of gaps between the model output and human translations for real-world applications, including suitability for post-editing.
\end{abstract}

\section{Introduction}

\begin{figure}[t]
    	\centering\includegraphics[width=\linewidth, frame]{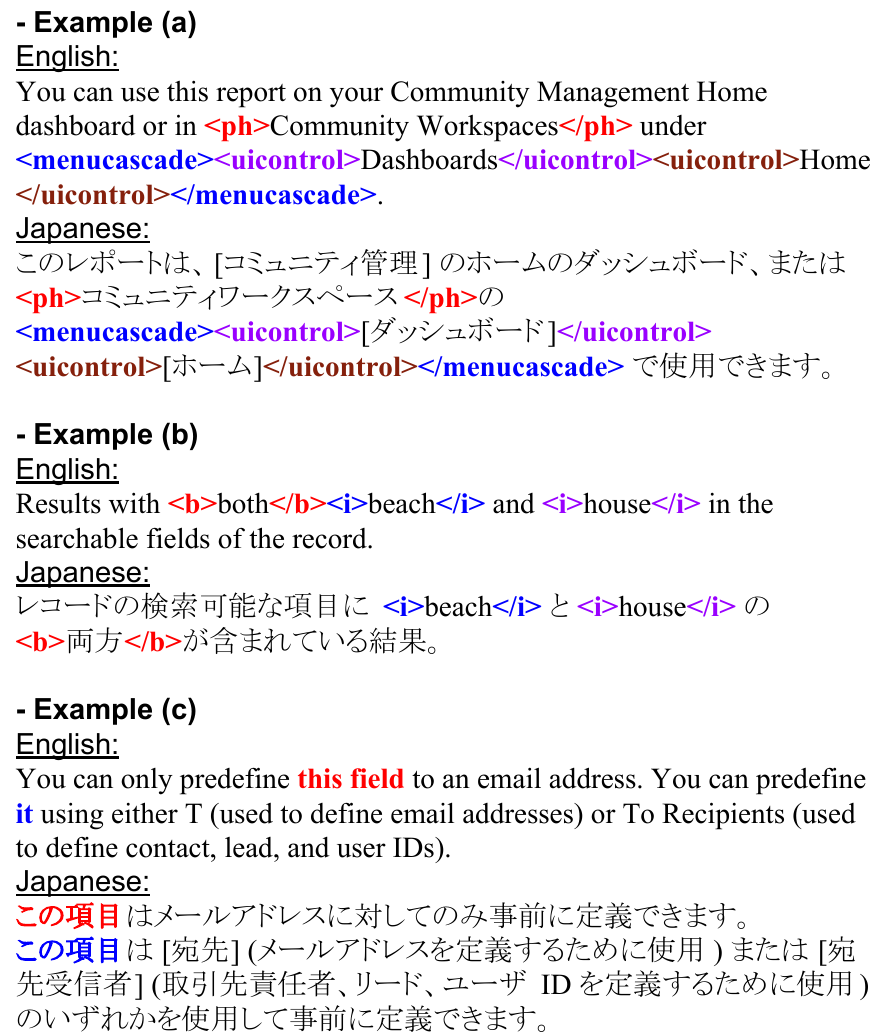}
\caption{English-Japanese examples in our dataset.}
\label{fig:examples}
\vspace{-4.0mm}
\end{figure}

Machine translation is a fundamental research area in the field of natural language processing (NLP).
To build a machine learning-based translation system, we usually need a large amount of bilingually-aligned text segments.
Examples of widely-used datasets are those included in WMT~\citep{wmt18} and LDC,\footnote{\url{https://www.ldc.upenn.edu/}} while new evaluation datasets are being actively created~\citep{noisy,contextMT,pronounMT}.
These existing datasets have mainly focused on translating plain text.

On the other hand, text data, especially on the Web, is not always stored as plain text, but often wrapped with markup languages to incorporate document structure and metadata such as formatting information.
Many companies and software platforms provide online help as Web documents, often translated into different languages to deliver useful information to people in different countries.
Translating such Web-structured text is a major component of the process by which companies localize their software or services for new markets, and human professionals typically perform the translation with the help of a {\it translation memory}~\citep{tm} to increase efficiency and maintain consistent terminology.
Explicitly handling such structured text can help bring the benefits of state-of-the-art machine translation models to additional real-world applications.
For example, structure-sensitive machine translation models may help human translators accelerate the localization process.

To encourage and advance research on translation of structured text, we collect parallel text segments from the public online documentation of a major enterprise software platform, while preserving the original XML structures.
%





In experiments, we provide baseline results for seven translation pairs from English, and one non-English pair.
We use standard neural machine translation (NMT) models, and additionally propose an XML-constrained beam search and several discrete copy mechanisms to provide solid baselines for our new dataset.
The constrained beam search contributes to accurately generating source-conditioned XML structures.
Besides the widely-used BLEU~\citep{bleu} scores, we also investigate more focused evaluation metrics to measure the effectiveness of our proposed methods.
In particular, we discuss trade-offs of using the copy mechanisms by focusing on translation of named entities and numerical words.
We further report detailed human evaluation and analysis to understand what is already achieved and what needs to be improved for the purpose of helping the human translators (a post-editing context).
As our dataset represents a single, well-defined domain, it can also serve as a corpus for domain adaptation research (either as a source or target domain).
We release our dataset publicly, and discuss potential for future expansion in Section~\ref{sec:related}.

\section{Collecting Data from Online Help}
This section describes how we constructed our new dataset for XML-structured text translation.

\paragraph{Why high quality?}
We start from the publicly-available online help of a major international enterprise software-as-a-service (SaaS) platform.
The software is provided in many different languages, and its multilingual online documentation has been localized and maintained for 15 years by the same localization service provider and in-house localization program managers.
Since the beginning they have been storing translations in a translation memory (i.e. computer-assisted translation tool) to increase quality and terminology consistency.
The documentation makes frequent use of structured formatting (using XML) to convey information to readers, so the translators have aimed to ensure consistency of formatting and markup structure, not just text content, between languages.

\paragraph{How many languages?}
The web documentation currently covers 16 non-English languages translated from English.
These 16 languages are Brazilian Portuguese, Danish, Dutch, Finnish, French, German, Italian, Japanese, Korean, Mexican Spanish, Norwegian, Russian, Simplified Chinese, Spanish, Swedish, and Traditional Chinese.
In practice, the human translation has been done from  English to the other languages, but all the languages could be potentially considered as both source and target because they contain the same tagging structure.

\subsection{Bilingual Web Page Alignments}
\label{subsec:page_align}

In this paper, we focus on each language pair separately, as an initial construction of our dataset.
Each page of the online documentation in the different languages is already aligned in the following two ways:

\begin{figure*}[t]
	\begin{center}
    	\includegraphics[width=\linewidth]{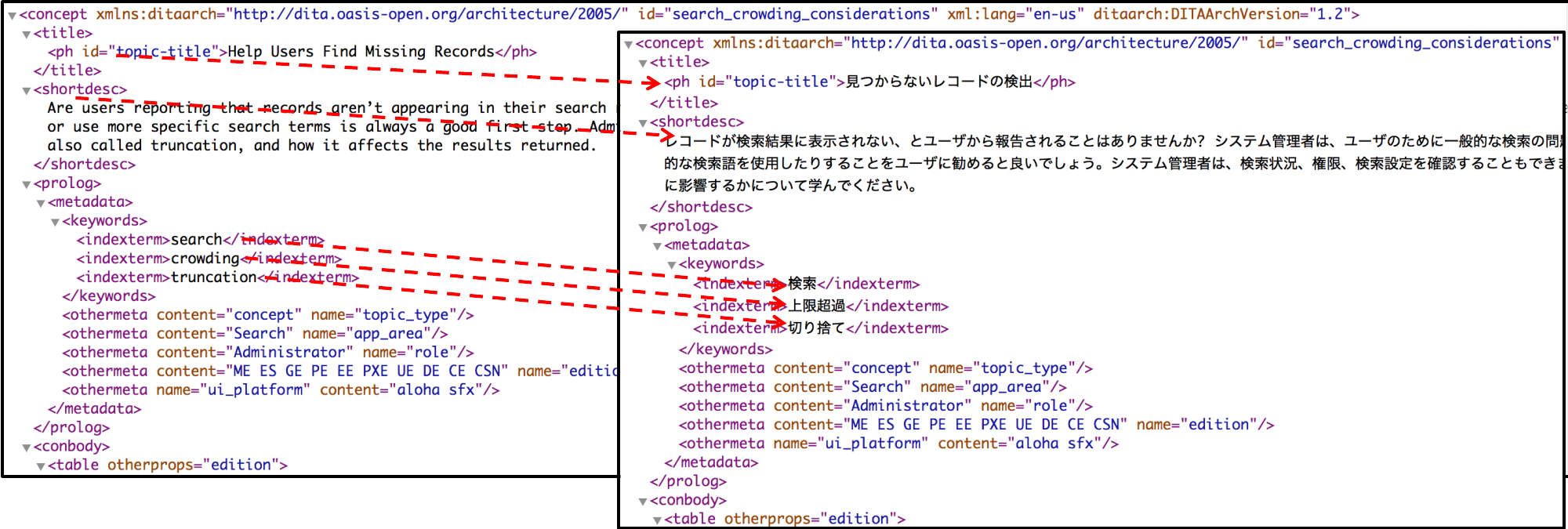}
    \end{center}
\caption{An aligned pair of English and Japanese XML files.}
\label{fig:xml_example}
\end{figure*}

\noindent
{\bf --} first, the same page has the same file name between languages; for example, if we have a page about ``WMT'', there would be {\tt /English/wmt.xml} and {\tt /Japanese/wmt.xml}, and

\noindent
{\bf --} second, most of the high-level XML elements are already aligned, because the original English files have been translated by preserving the same XML structures as much as possible in the localization process, to show the same content with the same formatting.
Figure~\ref{fig:xml_example} shows a typical pair of files and the alignment of their high-level XML elements.

Our dataset contains about 7,000 pairs of XML files for each language pair; for example, there are 7,336 aligned files for English-\{French, German, Japanese\}, 7,160 for English-\{Finnish, Russian\}, and 7,927 for Finnish-Japanese.\footnote{Some documents are not present, or not aligned, in all languages.}

\subsection{Extracting Parallel Text Segments}
\label{subsec:extract}

\paragraph{XML parsing and alignment}
For each language pair, we extract parallel text segments from XML structures.
We use the {\tt etree} module in a Python library called {\tt lxml}\footnote{\url{https://lxml.de/}} to process XML strings in the XML files.
Since the XML elements are well formed and translators keep the same tagging structure as much as their languages allow it, as described in Section~\ref{subsec:page_align}, we first linearize an XML-parsed file into a sequence of XML elements.
We then use a pairwise sequence alignment algorithm for each bilingually-aligned file, based on XML tag matching.
As a result, we have a set of aligned XML elements for the language pair.

\paragraph{Tag categorization}
Next, we manually define which XML elements should be translated, based on the following three categories:

\noindent
-- Translatable:

\noindent
A translatable tag (e.g. {\tt p}, {\tt xref}, {\tt note}) requires us to translate text inside the tag, and we extract translation pairs from this category.
In general, the translatable tags correspond to standalone text, and are thus easy to align in the sequence alignment step.

\noindent
-- Transparent:

\noindent
By contrast, a transparent tag (e.g. {\tt b}, {\tt ph}) is a formatting directive embedded as a child element in a translatable tag, and is not always well aligned due to grammatical differences among languages.
We keep the transparent tags embedded in the translatable tags.

\noindent
-- Untranslatable:

\noindent
In the case of untranslatable tags (e.g. {\tt sup}), we remove the elements.
The complete list of tag categorizations can be found in the supplementary material.

\begin{figure}[t]
	\begin{center}
    	\includegraphics[width=\linewidth]{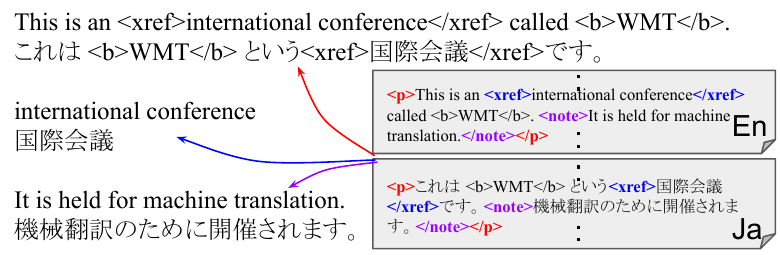}
    \end{center}
\caption{Extracting parallel text segments from aligned XML elements.}
\label{fig:extraction}
\end{figure}

\paragraph{Text alignment}
Figure~\ref{fig:extraction} shows how to extract parallel text segments based on the tag categorization.
There are three aligned translatable tags, and they result in three separate translation pairs.
The {\tt note} tag is translatable, so the entire element is removed when extracting the translation pair of the {\tt p} tag.
However, we do not remove nested translatable tags (like the {\tt xref} tag in this figure) when their {\it tail}\footnote{For example, the tail of the {\tt xref} tag in the English example corresponds to the word ``called.''} has text, to avoid missing phrases within sentences.
Next, we remove the root tag from each translation pair, because the correspondence is obvious.
We also remove fine-grained information such as attributes in the XML tags for the dataset; from the viewpoint of real-world usage, we can recover (or copy) the missing information as a post-processing step.
As a result of this process, a translation pair can consist of multiple sentences as shown in Example~(c) of Figure~\ref{fig:examples}.
We do not split them into single sentences, considering a recent trend of context-sensitive machine translation~\citep{contextMT,pronounMT,docMT1,docMT2}.
One can use split sentences for training a model, but an important note is that there is no guarantee that all the internal sentences are perfectly aligned.
We note that this structure-based alignment process means we do not rely on statistical alignment models to construct our parallel datasets.\footnote{Using HTML structures has been proven effective in aligning parallel sentences from the Web~\citep{html-alignment}, whereas we can directly start from the parallel files.}

\paragraph{Filtering}
We only keep translation pairs whose XML tag sets are consistent in both language sides, but we do not constrain the order of the tags to allow grammatical differences that result in tag reordering.
We remove duplicate translation pairs based on exact matching, and separate two sets of 2,000 examples each for development and test sets.
There are many possible experimental settings, and in this paper we report experimental results for seven English-based pairs, English-to-\{Dutch, Finnish, French, German, Japanese, Russian, Simplified Chinese\}, and one non-English pair, Finnish-to-Japanese.
The dataset thus provides opportunities to focus on arbitrary pairs of the 17 languages.
For each of the possible pairs, the number of training examples (aligned segments) is around 100,000.

\subsection{Detailed Dataset Statistics}

\begin{table}[t]
  \begin{center}
{\small
    \begin{tabular}{l|c|c}
    Language pair & Training data & Aligned files \\ \hline
    English- & \\
    ~~~~Dutch    & 100,756   & 7,160 \\
    ~~~~Finnish  & ~~99,759  & 7,160 \\
    ~~~~French   & 103,533   & 7,336 \\
    ~~~~German   & 103,247   & 7,336 \\
    ~~~~Japanese & 101,480   & 7,336 \\
    ~~~~Russian  & 100,332   & 7,160 \\
    ~~~~Simplified Chinese & ~~99,021 & 7,160 \\ \hdashline
    Finnish-Japanese    & 101,527 & 7,927 \\ \hline
    \end{tabular}
}
    \caption{The number of the translation examples in the training data used in our experiments.}
    \label{tb:stat}
  \end{center}

\end{table}

\begin{figure}[t]
	\begin{center}
    	\includegraphics[width=55mm]{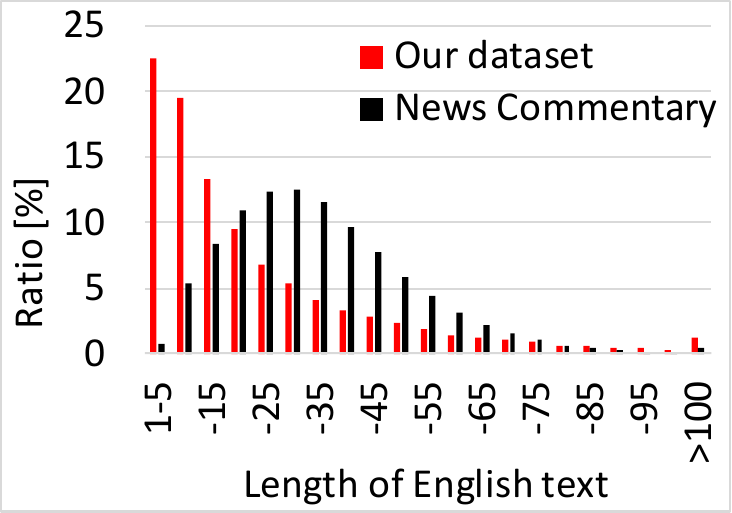}
    \end{center}
\caption{The length statistics of the English text in our English-French and the News Commentary datasets.}
\label{fig:len_stat}

	\begin{center}
    	\includegraphics[width=45mm]{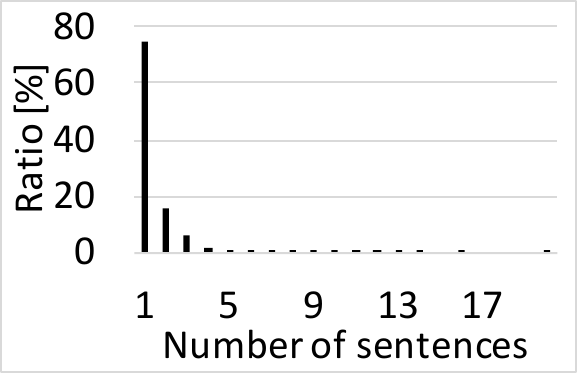}
    \end{center}
\caption{The statistics of the number of English sentences in the English-French translation pairs.}
\label{fig:sent_stat}

	\begin{center}
    	\includegraphics[width=45mm]{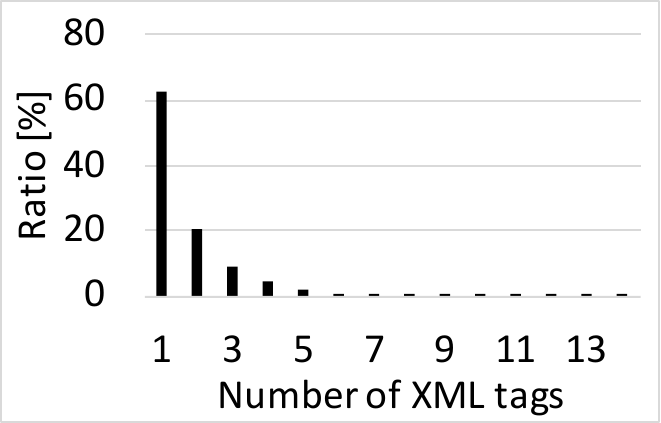}
    \end{center}
\caption{The statistics of the number of XML tags inside the English-French translation pairs.}
\label{fig:xml_tag_stat}
\end{figure}

Table~\ref{tb:stat} and Figure~\ref{fig:len_stat}, \ref{fig:sent_stat}, \ref{fig:xml_tag_stat} show more details about the dataset statistics.
We take our English-French dataset to show some detailed statistics, but the others also have the consistent statistics because all the pairs are grounded in the same English files.

\paragraph{Text lengths}
Due to the XML tag-based extraction, our dataset includes word- and phrase-level translations as well as sentence- and paragraph-level translations, and we can see in Figure~\ref{fig:len_stat} that there are many short text segments.
This is, for example, different from the statistics of the widely-used News Commentary dataset.
The text length is defined based on the number of subword tokens, following our experimental setting described below.

\paragraph{Sentence counts}
Another characteristic of our dataset is that the translation pairs can consist of multiple sentences, and Figure~\ref{fig:sent_stat} shows the statistics of the number of English sentences in the English-French translation pairs.
The number of sentences is determined with the sentence splitter from the Stanford CoreNLP toolkit~\citep{stanford}.

\paragraph{XML-tag counts}
As we remove the root tags from the XML elements in our dataset construction process, not all the text segments have XML tags inside them.
More concretely, about 25.5\% of the translation pairs have at least one internal XML tag, and Figure~\ref{fig:xml_tag_stat} shows the statistics.
For example, Example~(a) in Figure~\ref{fig:examples} has four XML tags, and Example~(b) has three.

\subsection{Evaluation Metrics}
We consider multiple evaluation metrics for the new dataset.
For evaluation, we use the {\it true-cased} and {\it detokenized} text, because our dataset is designed for an end-user, raw-document setting.

\paragraph{BLEU without XML}
We include the most widely-used metric, BLEU, without XML tags.
That is, we remove all the XML tags covered by our dataset and then evaluate BLEU.
The metric is compatible with the case where we use the dataset for plain text translation without XML.
To compute the BLEU scores, we use language-specific tokenizers; for example, we use Kytea~\citep{kytea} for Simplified Chinese and Japanese, and the Moses~\citep{moses} tokenizer for English, Dutch, Finnish, French, German, and Russian.

\paragraph{Named entities and numbers}
The online help frequently mentions named entities such as product names and numbers, and accurate translations of them are crucial for users.
Frequently, they are not translated but simply copied as English forms.
We evaluate corpus-level precision and recall for translation of the named entities and numerical tokens.
To extract the named entities and numerical words, we use a rule-based regex script, based on our manual analysis on our dataset.
The numerical words are extracted by
\begin{itemize}
\item[] {``[0-9.,\textbackslash'/:]*[0-9]+[0-9.,\textbackslash'/:]*''}.
\end{itemize}
The named entities are defined as
\begin{itemize}
\item[] {``[.,\textbackslash'/:a-zA-Z\$]*[A-Z]+[.,\textbackslash'/:a-zA-Z\$]*''}
\end{itemize}
appearing in a non-alphabetic language, Japanese, because in our dataset we observe that the alphabetic words in such non-alphabetic languages correspond to product names, country names, function names, etc.

\paragraph{XML accuracy, matching, and BLEU}
For each output text segment, we use the {\tt etree} module to check if it is a valid XML structure by wrapping it with a dummy root node.
Then the XML accuracy score is the number of the valid outputs, divided by the number of the total evaluation examples.
We further evaluate how many translation outputs have exactly the same XML structures as their corresponding reference text (an XML matching score).
If a translation output matches its reference XML structure, both the translation and reference are split by the XML tags.
We then evaluate corpus-level BLEU by comparing each split segment one by one.
If an output does not match its reference XML structure, the output is treated as empty to penalize the irrelevant outputs.

\section{Machine Translation with XML Tags}

We use NMT models to provide competitive baselines for our dataset.
This section first describes how to handle our dataset with a sequential NMT model.
We then propose a simple constrained beam search for accurately generating XML structures conditioned by source information.
We further incorporate multiple copy mechanisms to strengthen the baselines.

\subsection{Sequence-to-Sequence NMT}
\label{subsec:nmt}
The task in our dataset is to translate text with structured information, and therefore we consider using syntax-based NMT models.
A possible approach is incorporating parse trees or parsing algorithms into NMT models~\citep{eriguchi2016,eriguchi2017}, and another is using sequential models on linearized structures~\citep{roee2017}.
We employ the latter approach to incorporate source-side and target-side XML structures, and note that this allows using standard sequence-to-sequence models without modification.

We have a set of parallel text segments for a language pair $(\mathcal{X}, \mathcal{Y})$, and the task is translating a text segment $x\in\mathcal{X}$ to another $y\in\mathcal{Y}$.
Each $x$ in the dataset is represented with a sequence of tokens including some XML tags: $x=[x_1, x_2, \ldots, x_N]$, where $N$ is the length of the sequence.
Its corresponding reference $y$ is also represented with a sequence of tokens: $y=[y_1, y_2, \ldots, y_M]$, where $M$ is the sequence length.
Any tokenization method can be used, except that the XML tags should be individual tokens.

To learn translation from $x$ to $y$, we use a {\it transformer} model~\citep{transformer}.
In our $K$-layer transformer model, each source token $x_i$ in the $k$-th $(k\in[1,K])$ layer is represented with
\begin{eqnarray}
h^x_k(x_i) = f(i, h^x_{k-1})\in\mathbb{R}^{d},
\end{eqnarray}
where $i$ is the position information, $d$ is the dimensionality of the model, and $h^x_{k-1}=[h^x_{k-1}(x_1), h^x_{k-1}(x_2), \ldots, h^x_{k-1}(x_N)]$ is the sequence of the vector representations in the previous layer.
$h^x_0(x_i)$ is computed as $h^x_0(x_i) = \sqrt{d} \cdot v(x_i) + e(i)$,
where $v(x_i)\in\mathbb{R}^{d}$ is a token embedding, and $e(i)\in\mathbb{R}^{d}$ is a positional embedding.

Each target-side token $y_j$ is also represented in a similar way:
\begin{eqnarray}
h^y_k(y_j) = g(j, h^x_k, h^y_{k-1}) \in\mathbb{R}^{d},
\end{eqnarray}
where only $[h^y_{k-1}(y_1), h^y_{k-1}(y_2), \ldots, h^y_{k-1}(y_j)]$ is used from $h^y_{k-1}$.
In the same way as the source-side embeddings, $h^y_0(y_j)$ is computed as $h^y_0(y_j) = \sqrt{d} \cdot v(y_j) + e(j)$.
For more details about the parameterized functions $f$ and $g$, and the positional embeddings, please refer to \citet{transformer}.

Then $h^y_K(y_j)$ is used to predict the next token $w$ by a softmax layer: $p_g(w|x, y_{\leq j}) = \mathrm{softmax}(W h^y_K(y_j) + b)$,
where $W\in\mathbb{R}^{|\mathbb{V}|\times d}$ is a weight matrix, $b\in\mathbb{R}^{|\mathbb{V}|}$ is a bias vector, and $\mathbb{V}$ is the vocabulary.
The loss function is defined as follows:
\begin{eqnarray}
L(x, y) = -\sum^{M-1}_{j=1} \log{p_g(w=y_{j+1}|x, y_{\leq j})},
\end{eqnarray}
where we assume that $y_1$ is a special token ${\tt BOS}$ to indicate the beginning of the sequence, and $y_M$ is an end-of-sequence token ${\tt EOS}$.
Following \citet{tai_softmax} and \citet{tai_softmax_2}, we use $W$ as an embedding matrix, and we share the single vocabulary $\mathbb{V}$ for both $\mathcal{X}$ and $\mathcal{Y}$.
That is, each of $v(x_i)$ or $v(y_j)$ is equivalent to a row vector in $W$.

\subsection{XML-Constrained Beam Search}

At test time, standard sequence-to-sequence generation methods do not always output valid XML structures, and even if an output is a valid XML structure, it does not always match the tag set of its source-side XML structure.
To generate source-conditioned XML structures as accurately as possible, we propose a simple constrained beam search method.
We add three constrains to a standard beam search method.
First, we keep track of possible tags based on the source input, and allow the model to open only a tag that is present in the input and has not yet been covered.
Second, we keep track of the most recently opened tag, and allow the model to close the tag.
Third, we do not allow the model to output ${\tt EOS}$ before opening and closing all the tags used in the source sentence.
Algorithm~1 in the supplementary material shows a comprehensive pseudo code.


\subsection{Reformulating a Pointer Mechanism}
\label{subsec:copy}

We consider how to further improve our NMT system, by using multiple {\it discrete} copy mechanisms.
Since our dataset is based on XML-structured technical documents, we want our NMT system to copy (A) relevant text segments in the target language if there are very similar segments in the training data, and (B) named entities (e.g. product names), XML tags, and numbers directly from the source. 
For the copy mechanisms, we follow the general idea of the {\it pointer} used in \citet{pointer}. 


For the sake of discrete decisions, we reformulate the pointer method.
Following the previous work, we have a sequence of tokens which are targets of our pointer method: $c=[c(z_1), c(z_2), \ldots, c(z_U)]$,
where $c(z_i)\in\mathbb{R}^{d}$ is a vector representation of the $i$-th token $z_i$, and $U$ is the sequence length.
As in Section~\ref{subsec:nmt}, we have $h^y_K(y_j)$ to predict the $(j+1)$-th token.
Before defining an attention mechanism between $h^y_K(y_j)$ and $c$, we append a parameterized vector $c(z_0)=c'$ to $c$.
We expect $c'$ to be responsible for decisions of ``not copying'' tokens, and the idea is inspired by adding a ``null'' token in natural language inference~\citep{null}.

We then define attention scores between $h^y_K(y_j)$ and the expanded $c$: $a(j, i)=score(h^y_K(y_j), c_i, c)$,
where the normalized scoring function $score$ is implemented as a single-head attention model proposed in \citet{transformer}.
If the next reference token $y_{j+1}$ is not included in the copy target sequence, the loss function is defined as follows:
\begin{eqnarray}
L(x, y_{\leq j}, c) = -\log{a(j, 0)},
\end{eqnarray}
and otherwise the loss function is as follows:
\begin{eqnarray}
L(x, y_{\leq j}, c) = -\log{\sum_{i,~\mathrm{s.t.}~z_i=y_{j+1}} a(j, i)},
\end{eqnarray}
and then the total loss function is $L(x, y) + \sum_{j=1}^{M-1} L(x, y_{\leq j}, c)$.
The loss function solely relies on the cross-entropy loss for single probability distributions, whereas the pointer mechanism in \citet{pointer} defines the cross-entropy loss for weighted summation of multiple distributions.

At test time, we employ a discrete decision strategy for copying tokens or not.
More concretely, the output distribution is computed as
\begin{eqnarray}
\delta \cdot p_g(w|x, y_{\leq j}) + (1-\delta) \cdot p_c(w|x, y_{\leq j}),
\end{eqnarray}
where $p_c(w|x, y_{\leq j})$ is computed by aggregating $[a(j, 1), \ldots, a(j, U)]$.
$\delta$ is 1 if $a(j, 0)$ is the largest among $[a(j, 0), \ldots, a(j, U)]$, and otherwise $\delta$ is 0.

\paragraph{Copy from Retrieved Translation Pairs}
\citet{jiatao} presented a retrieval-based NMT model, based on the idea of translation memory~\citep{tm}.
Following \citet{jiatao}, we retrieve the most relevant translation pair $(x', y')$ for each source text $x$ in the dataset. 
In this case, we set $[z_1, \ldots, z_U]=[y'_2, \ldots, y'_{M'}]$ and $c=[h^y_{K}(y'_1), \ldots, h^y_{K}(y'_{M'-1})]$,
where $M'$ is the length of $y'$, and each vector in $c$ is computed by the same transformer model in Section~\ref{subsec:nmt}.
For this retrieval copy mechanism, we denote $p_c$ and $\delta$ as $p_r$ and $\delta_r$, respectively.

\paragraph{Copy from Source Text}
To allow our NMT model to directly copy certain tokens from the source text $x$ when necessary, we follow \citet{pointer}.
We set $[z_1, \ldots, z_U]=[x_1, \ldots, x_{N}]$ and $c=[h^x_{K}(x_1), \ldots, h^x_{K}(x_{N})]$,
and we denote $p_c$ and $\delta$ as $p_s$ and $\delta_s$, respectively.

We have the single vocabulary $\mathbb{V}$ to handle all the tokens in both languages $\mathcal{X}$ and $\mathcal{Y}$, and we can combine the three output distributions at each time step in the text generation process:
\begin{eqnarray}
\label{eq:final_p}
(1-\delta_s) p_s + \delta_s (\delta_r p_g + (1-\delta_r) p_r).
\end{eqnarray}
The copy mechanism is similar to the multi-pointer-generator method in \citet{deca}, but our method employs rule-based discrete decisions.
Equation~(\ref{eq:final_p}) first decides whether the NMT model copies a source token.
If not, our method then decides whether the model copies a retrieved token.

\section{Experimental Settings}

This section describes our experimental settings.
More details are described in the supplementary material.

\subsection{Tokenization and Detokenization}

We used the SentencePiece toolkit~\citep{spm} for sub-word tokenization and detokenization for the NMT outputs.

\paragraph{Without XML tags}
If we remove all the XML tags from our dataset, the task becomes a plain MT task.
We carried out our baseline experiments for the plain text translation task, and for each language pair we trained a joint SentencePiece model to obtain its shared sub-word vocabulary.
For training each NMT model, we used training examples whose maximum token length is 100.

\paragraph{With XML tags}
For our XML-based experiments, we also trained a joint SentencePiece model for each language pair, where one important note is that all the XML tags are treated as user-defined special tokens in the toolkit.
This allows us to easily implement the XML-constrained beam search.
We also set the three tokens {\tt \&amp;}, {\tt \&lt;}, and {\tt \&gt;} as special tokens.

\subsection{Model Configurations}
We implemented the transformer model with $K=6$ and $d=256$ as a competitive baseline model.
We trained three models for each language pair:
\begin{itemize}
\item[] ``OT'' (trained only with text without XML),
\item[] ``X'' (trained with XML), and
\item[] ``X$_\mathrm{rs}$'' (XML and the copy mechanisms).
\end{itemize}
For each setting, we tuned the model on the development set and selected the best-performing model in terms of BLEU scores {\it without} XML, to make the tuning process consistent across all the settings.

\section{Results}

\begin{table*}[t]
  \begin{center}
{\small
    \begin{tabular}{l|c|c|c|c|c|c|c|c}
    & & NE\&NUM & & NE\&NUM & & NE\&NUM & & NE\&NUM  \\
	& BLEU & Precision, Recall & BLEU & Precision, Recall & BLEU & Precision, Recall & BLEU & Precision, Recall  \\ \hline

	& \multicolumn{2}{c|}{English-to-Japanese} & \multicolumn{2}{c|}{English-to-Chinese} & \multicolumn{2}{c|}{English-to-French} & \multicolumn{2}{c}{English-to-German} \\ \hdashline
     OT              & 61.61 & 89.84, 89.84 & 58.06 & 94.91, 93.62 & 64.07 & 88.64, 85.64 & 50.51 & 88.40, 86.55 \\
     X               & 62.00 & 92.54, 90.51 & 58.61 & 94.56, 93.44 & 63.98 & 87.48, 86.98 & 50.96 & 88.79, 86.43 \\
     X$_\mathrm{rs}$ & 64.25 & 91.64, 90.98 & 60.05 & 94.44, 94.27 & 63.51 & 88.42, 85.64 & 52.91 & 88.00, 86.78 \\ \hdashline
     X$^{\mathrm{(T)}}_\mathrm{rs}$
                     & 64.34 & 93.39, 91.75 & 59.86 & 93.49, 93.11 & 65.04 & 88.98, 88.31 & 52.69 & 88.22, 88.45 \\ \hline
    
	& \multicolumn{2}{c|}{English-to-Finnish} & \multicolumn{2}{c|}{English-to-Dutch} & \multicolumn{2}{c|}{English-to-Russian} & \multicolumn{2}{c}{Finnish-to-Japanese} \\ \hdashline
     OT              & 43.97 & 87.58, 84.99 & 59.54 & 90.89, 88.59 & 43.28 & 89.67, 85.26 & 54.55 & 90.45, 89.69  \\
     X               & 42.84 & 83.17, 85.55 & 60.18 & 90.41, 90.26 & 43.44 & 87.96, 88.35 & 54.69 & 93.47, 89.29 \\
     X$_\mathrm{rs}$ & 45.10 & 86.41, 86.49 & 60.58 & 88.76, 90.11 & 46.73 & 88.65, 89.55 & 57.92 & 93.02, 89.03 \\ \hdashline
     X$^{\mathrm{(T)}}_\mathrm{rs}$
                     & 45.71 & 87.38, 88.91 & 61.01 & 87.66, 90.84 & 46.44 & 86.90, 89.59 & 57.06 & 93.39, 89.38 \\ \hline
     
    \end{tabular}
}
    \caption{Automatic evaluation results {\it without} XML on the development set, and the test set for X$_\mathrm{rs}$.}
    \label{tb:noXML}
  \end{center}

\end{table*}

\begin{table}[t]
  \begin{center}
{\small
    \begin{tabular}{l|c|c}
    Training data    & Our dev set & newstest2014 \\ \hline
    Our dataset (no XML)  & 64.07 & ~~7.35  \\
    w/ 10K news & 63.66 & 14.02  \\
    w/ 20K news & 64.31 & 16.30 \\ \hdashline
    Only 10K news & ~~0.90 & ~~2.66 \\
    Only 20K news & ~~2.35 & ~~6.72 \\ \hline
    
    \end{tabular}
}
    \caption{Domain adaptation results (BLEU). The models are tuned on our development set.}
    \label{tb:da}
  \end{center}

\end{table}

\begin{table*}[t]
  \begin{center}
{\small
    \begin{tabular}{l|c|c|c|c|c|c|c|c}
    & & XML & & XML & & XML & & XML  \\
	& BLEU & Acc., Match & BLEU & Acc., Match & BLEU & Acc., Match & BLEU & Acc., Match  \\ \hline

	& \multicolumn{2}{c|}{English-to-Japanese} & \multicolumn{2}{c|}{English-to-Chinese} & \multicolumn{2}{c|}{English-to-French} & \multicolumn{2}{c}{English-to-German} \\ \hdashline
     X               & 59.77 & 99.80, 99.55 & 57.01 & 99.95, 99.70 & 61.81 & 99.60, 99.30 & 48.91 & 99.85, 99.25   \\
     X$_\mathrm{rs}$ & 62.06 & 99.80, 99.40 & 58.43 & 99.90, 99.60 & 61.87 & 99.80, 99.50 & 51.16 & 99.75, 99.30  \\ \hdashline
     X$^{\mathrm{(T)}}_\mathrm{rs}$
                     & 62.27 & 99.95, 99.60 & 57.92 & 99.75, 99.40 & 63.19 & 99.80, 99.35 & 50.47 & 99.80, 99.20 \\ \hline
    
	& \multicolumn{2}{c|}{English-to-Finnish} & \multicolumn{2}{c|}{English-to-Dutch} & \multicolumn{2}{c|}{English-to-Russian} & \multicolumn{2}{c}{Finnish-to-Japanese} \\ \hdashline
     X               & 41.98 & 99.65, 99.25 & 57.86 & 99.60, 99.25 & 40.72 & 99.60, 98.95 & 52.14 & 99.90, 99.30 \\
     X$_\mathrm{rs}$ & 43.57 & 99.50, 99.25 & 58.51 & 99.70, 99.30 & 44.42 & 99.75, 99.25 & 55.20 & 99.65, 98.90 \\ \hdashline
     X$^{\mathrm{(T)}}_\mathrm{rs}$
                     & 44.22 & 99.90, 99.65 & 60.19 & 99.90, 99.85 & 44.25 & 99.80, 99.35 & 54.05 & 99.60, 98.75 \\ \hline

    \end{tabular}
}
    \caption{Automatic evaluation results {\it with} XML on the development set, and the test set for X$_\mathrm{rs}$.}
    \label{tb:XML}
  \end{center}

\end{table*}

Table~\ref{tb:noXML} and \ref{tb:XML} show the detailed results on our development set, and for the X$_\mathrm{rs}$ model, we also show the results (X$^{\mathrm{(T)}}_\mathrm{rs}$) on our test set to show our baseline scores for future comparisons.
Simplified Chinese is written as ``Chinese'' in this section.

\subsection{Evaluation without XML}
\label{subsec:noXML}

We first focus on the two evaluation metrics: BLEU without XML, and named entities and numbers (NE\&NUM).
In Table~\ref{tb:noXML}, a general observation from the comparison of OT and X is that including segment-internal XML tags tends to improve the BLEU scores.
This is not surprising because the XML tags provide information about explicit or implicit alignments of phrases.
However, the BLEU score of the English-to-Finnish task significantly drops, which indicates that for some languages it is not easy to handle tags within the text.

Another observation is that X$_\mathrm{rs}$ achieves the best BLEU scores, except for English-to-French.
In our experiments, we have found that the improvement of BLEU comes from the retrieval method, but it degrades the NE\&NUM scores, especially the precision.
Then copying from the source tends to recover the NE\&NUM scores, especially for the recall.
We have also observed that using beam search, which improves BLEU scores, degrades the NE\&NUM scores.
A lesson learned from these results is that work to improve BLEU scores can sometimes lead to degradation of other important metrics.

\paragraph{Compatibility with other domains}
Our dataset is limited to the domain of online help, but we can use it as a seed corpus for domain adaptation if our dataset contains enough information to learn basic grammar translation.
We conducted a simple domain adaptation experiment in English-to-French by adding 10,000 or 20,000 training examples of the widely-used News Commentary corpus.
We used the newstest2014 dataset for evaluation in the news domain.
From Table~\ref{tb:da}, we can see that a small amount of the news-domain data significantly improves the target-domain score, and we expect that our dataset plays a good role in domain adaptation for all the covered 17 languages.

\subsection{Evaluation with XML}

\begin{table}[t]
  \begin{center}
{\small
    \begin{tabular}{l|c|c}
            &      & XML \\
            & BLEU & Acc., Match \\ \hline
    w/ XML constraint  & 59.77 & 99.80, 99.55 \\
    w/o XML constraint & 58.02 & 98.70, 98.10 \\ \hline

    \end{tabular}
}
    \caption{Effects of the XML-constrained beam search.}
    \label{tb:xml_search}
  \end{center}

  \begin{center}
{\small
    \begin{tabular}{l|r}

     & Count \\ \hline
    Copied from source text & 1,638 \\
    Copied from retrieved translation & 24 \\
    Generated from vocabulary & 11 \\ \hline

    \end{tabular}
}
    \caption{Statistics of the generated XML tags.}
    \label{tb:tag_copy}
  \end{center}

\end{table}

\begin{figure*}[t]
	\begin{center}
    	\includegraphics[width=\linewidth, frame]{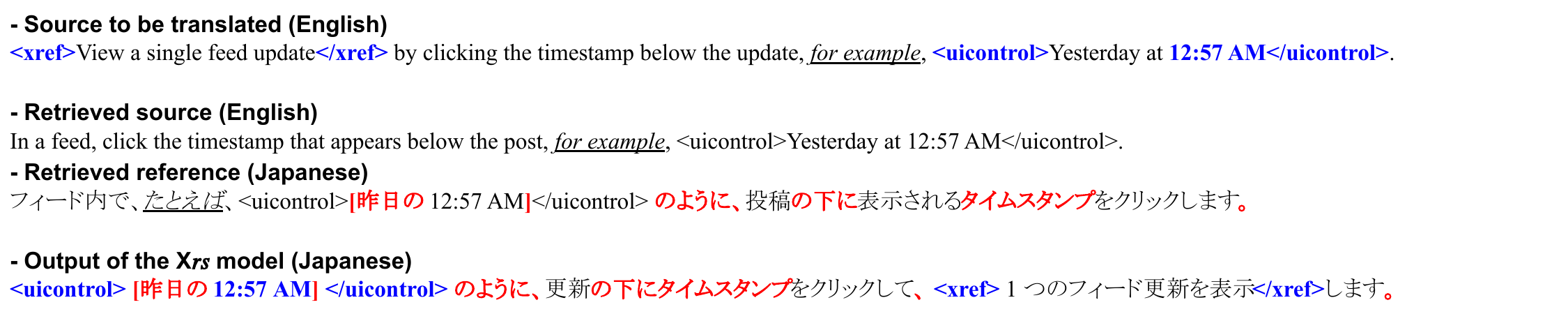}
    \end{center}
\caption{An example of the translation results of the X$_\mathrm{rs}$ model on the English-Japanese test set.}
\label{fig:trans_result}
\end{figure*}

Table~\ref{tb:XML} shows the evaluation results with XML.
Again, we can see that X$_\mathrm{rs}$ performs the best in terms of the XML-based BLEU scores, but the absolute values are lower than those in Table~\ref{tb:noXML} due to the more rigid segment-by-segment comparisons.
This table also shows that the XML accuracy and matching scores are higher than 99\% in most of the cases.
Ideally, the scores could be 100\%, but in reality, we set the maximum length of the translations; as a result, sometimes the model cannot find a good path within the length limitation.
Table~\ref{tb:xml_search} shows how effective our method is, based on the English-to-Japanese result, and we observed the consistent trend across the different languages.
These results show that our method can accurately generate the relevant XML structures.

\paragraph{How to recover XML attributes?}
As described in Section~\ref{subsec:extract}, we removed all the attributes from the original XML elements for simplicity.
However, we need to recover the attributes when we use our NMT model in the real-world application.
We consider recovering the XML attributes by the copy mechanism from the source; that is, we can copy the attributes from the XML elements in the original source text, if the XML tags are copied from the source.
Table~\ref{tb:tag_copy} summarizes how our model generates the XML tags on the English-Japanese development set.
We can see in the table that most of the XML tags are actually copied from the source.

Figure~\ref{fig:trans_result} shows an example of the output of the X$_\mathrm{rs}$ model.
For this visualization, we merged all the subword tokens to form the standard words.
The tokens in blue are explicitly copied from the source, and we can see that the time expression ``12:57 AM'' and the XML tags are copied as expected.
The output also copies some relevant text segments (in red) from the retrieved translation.
Like this, we can explicitly know which words are copied from which parts, by using our multiple discrete copy mechanisms.
One surprising observation is that the underlined phrase ``for example'' is missing in the translation result, even though the BLEU scores are higher than those on other standard public datasets.
This is a typical error called {\it under translation}.
Therefore, no matter how large the BLEU scores are, we definitely need human corrections (or post editing) before providing the translation results to customers.

\subsection{Human Evaluation by Professionals}

One important application of our NMT models is to help human translators; translating online help has to be precise, and thus any incomplete translations need post-editing.
We asked professional translators at a vendor to evaluate our test set results (with XML) for the English-to-\{Finnish, French, German, Japanese\} tasks.
For each language pair, we randomly selected 500 test examples, and every example is given an integer score in [1, 4].
A translation result is rated as ``4'' if it can be used without any modifications, ``3'' if it needs simple post-edits, ``2'' if it needs more post-edits but is better than nothing, and ``1'' if using it is not better than translating from scratch.

\begin{figure}[t]
	\begin{center}
    	\includegraphics[width=50mm]{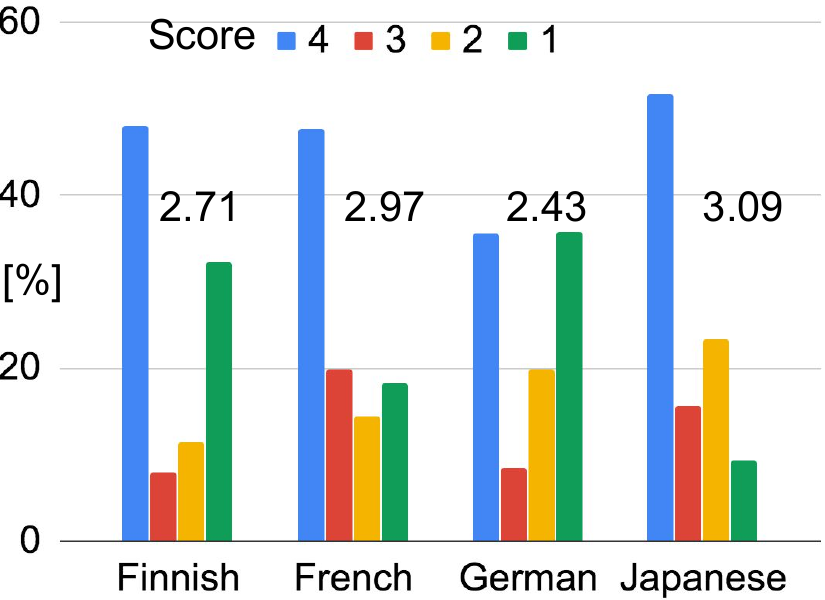}
    \end{center}
\caption{Human evaluation results for the X$_\mathrm{rs}$ model. ``4'' is the best score, and ``1'' is the worst.}
\label{fig:human_eval}
\end{figure}

Figure~\ref{fig:human_eval} shows the summary of the evaluation to see the ratio of each score, and the average scores are also shown.
A positive observation for all the four languages is that more than 50\% of the translation results are evaluated as complete or useful in post-editing.
However, there are still many low-quality translation results; for example, around 30\% of the Finnish and German results are evaluated as useless.
Moreover, the German results have fewer scores of ``4'', and it took 12 hours for the translators to evaluate the German results, whereas it took 10 hours for the other three languages.
To further make our NMT models useful for post-editing, we have to improve the translations scored as ``1''.

\paragraph{Detailed error analysis}
We also asked the translators to note what kinds of errors exist for each of the evaluated examples.
All the errors are classified into the six types shown in Table~\ref{tb:error}, and each example can have multiple errors.
The ``Formatting'' type is our task-specific one to evaluate whether the XML tags are correctly inserted.
We can see that the Finnish results have significantly more XML-formatting errors, and this result agrees with our finding that handling the XML tags in Finnish is harder than in other languages, as discussed in Section~\ref{subsec:noXML}.
It is worth further investigating such language-specific problems.

The ``Accuracy'' type covers major issues of NMT, such as adding irrelevant words, skipping important words, and mistranslating phrases.
As discussed in previous work~\citep{rep}, reducing the typical errors covered by the ``Accuracy'' type is crucial.
We have also noticed that the NMT-specific errors would slow down the human evaluation process, because the NMT errors are different from translation errors made by humans.
The other types of errors would be reduced by improving language models, if we have access to in-domain monolingual corpora.

\paragraph{Can MT help the localization process?}
In general, it is encouraging to observe many ``4'' scores in Figure~\ref{fig:human_eval}.
However, one important note is that it takes significant amount of time for the translators to verify the NMT outputs are good enough.
That is, having better scored NMT outputs does not necessarily lead to improving the productivity of the translators; in other words, we need to take into account the time for the quality verification when we consider using our NMT system for that purpose.
Previous work has investigated the effectiveness of NMT models for post-editing~\citep{smt_or_nmt_pe}, but it has not yet been investigated whether using NMT models can improve the translators' productivity alongside the use of a well-constructed translation memory~\citep{tm}.
Therefore, our future work is investigating the effectiveness of using the NMT models in the real-world localization process where a translation memory is available.

\begin{table}[t]
  \begin{center}
{\small
    \begin{tabular}{l|r|r|r|r}
                & \multicolumn{1}{c|}{Finnish} & \multicolumn{1}{c|}{French} & \multicolumn{1}{c|}{German} & \multicolumn{1}{c}{Japanese} \\ \hline
    Accuracy    & 30.0 & 32.8 & 37.4 & 37.4 \\
    Readability & 20.6 & 20.4 &  0.8 & 17.4 \\
    Formatting  & 10.6 &  0.0 &  0.8 & 1.0 \\
    Grammar     & 20.2 & 10.0 & 11.4 & 5.8 \\
    Structure   & 10.2 &  2.8 &  2.0 & 1.2 \\
    Terminology & 12.0 &  3.0 &  2.4 & 0.6 \\ \hline
    
    \end{tabular}
}
    \caption{Ratio [\%] of six error types.}
    \label{tb:error}
  \end{center}

\end{table}

\section{Related Work and Discussions}
\label{sec:related}
Automatic extraction of parallel sentences has a long history~\citep{align}, and usually statistical methods and dictionaries are used.
By contrast, our data collection solely relies on the XML structure, because the original data have been well structured and aligned.
Recently, collecting training corpora is the most important in training NLP models, and thus it is recommended to maintain well-aligned documents and structures when building multilingual online services.
That will significantly contribute to the research of language technologies.

We followed the syntax-based NMT models~\citep{eriguchi2016,eriguchi2017,roee2017} to handle the XML structures.
One significant difference between the syntax-based NMT and our task is that we need to output source-conditioned structures that are able to be parsed as XML, whereas the syntax-based NMT models do not always need to follow formal rules for their output structures.
In that sense, it would be interesting to relate our task to source code generation~\citep{codegen} in future work.

Our dataset has significant potential to be further expanded.
Following the context-sensitive translation~\citep{contextMT,pronounMT,docMT1,docMT2}, our dataset includes translations of multiple sentences.
However, the translatable XML tags are separated, so the page-level global information is missing.
One promising direction is thus to create page-level translation examples.
Finally, considering the recent focus on multilingual NMT models~\citep{zeroshot}, multilingually aligning the text will enrich our dataset.

\section{Conclusion}
We have presented our new dataset for XML-structured text translation.
Our dataset covers 17 languages each of which can be either source or target of machine translation.
The dataset is of high quality because it consists of professional translations for an online help domain.
Our experiments provide baseline results for the new task by using NMT models with an XML-constrained beam search and discrete copy mechanisms.
We further show detailed human analysis to encourage future research focusing on how to apply machine translation to help human translators in practice.

\section*{Acknowledgements}
We thank anonymous reviewers and Xi Victoria Lin for their helpful feedback.

\bibliography{bibtex}
\bibliographystyle{acl_natbib}

\appendix

\section*{Supplementary Material}

\section{Dataset Construction}

\subsection{XML Tag Categorization}
The three manually-categorized XML tags are as follows:

\noindent
{\bf -- translatable}
\{title, p, li, shortdesc, indexterm, note, section, entry, dt, dd, fn, cmd, xref, info, stepresult, stepxmp, example, context, term, choice, stentry, result, navtitle, linktext, postreq, prereq, cite, chentry, sli, choption, chdesc, choptionhd, chdeschd, sectiondiv, pd, pt, stepsection, index-see, conbody, fig, body, ul\},

\noindent
{\bf -- transparent}
\{ph, uicontrol, b, parmname, i, u, menucascade, image, userinput, codeph, systemoutput, filepath, varname, apiname\},

\noindent
{\bf -- untranslatable}
\{sup, codeblock, prodname\}.

\noindent
Among them, our pre-processed dataset has \{ph, xref, uicontrol, b, codeph, parmname, i, title, menucascade, varname, userinput, filepath, term, systemoutput, cite, li, ul, p, note, indexterm, u, fn\} embedded in the text as the actual XML tags.

\subsection{URL Normalization}
We have noticed that URLs are frequently mentioned in our dataset, and they are copied from one language to another.
For simplicity, we replaced URL-like strings with placeholders.
For example, the following sentence
\begin{itemize}
    \item[] ``http://aclweb.org/anthology/ has been moved to https://aclanthology.coli.uni-saarland.de/.''
\end{itemize}
is changed to
\begin{itemize}
    \item[] ``{\tt \#URL1\#} has been moved to {\tt \#URL2\#}.''
\end{itemize}
by keeping the correspondence between the same URLs in both sides of the paired languages.
The evaluation is performed with the URL-anonymized form of the text.

\section{XML-Constrained Beam Search}

Algorithm~\ref{alg:beam} shows a comprehensive pseudo code of our XML-constrained beam search.
$T$ is the set of possible XML tag types, $B$ is a beam size, and $L$ is a maximum length of the generated sequences.
Following \citet{wat_oda}, we use a length penalty $\alpha$.
The proposed beam search ensures a valid XML structure conditioned by its source information, unless the generated sequence does not violate the maximum length constraint.
It should be noted that this does not always lead to exactly the same structure as the structure of its reference text.

\begin{algorithm}[tp]
\caption{XML-constrained beam search}
\label{alg:beam}
{\small
\begin{algorithmic}[1]
\Function{ConstrainedBeamSearch}{$x$, $T$, $B$, $L$, $\alpha$}

\State $C = []$ \Comment{Candidates in the beam search}
\For{$i$ in $1\ldots B$}

\State $y=[{\tt BOS}]$ \Comment{Output token sequence}
\State $s = 0.0$ \Comment{Score}
\State $t = [t_1, \ldots, t_T]$ \Comment{Possible XML tag types in $x$}
\State $t' = []$ \Comment{History of opened tags}
\State $C$.append$(\{y, s, t, t'\})$

\EndFor

\State
\While{max length $< L$ and $C[0].y[-1]$ is not ${\tt EOS}$}

\For{$i$ in $1\ldots B$}
\If{$C[i].y[-1]$ is ${\tt EOS}$}
\State $\ell_i = C[i].s$

\Else
\State $\ell_i = \log{p(w|x, C[i].y)} \in\mathbb{R}^{|\mathbb{V}|}$
\State $\ell_i \mathrel{+}= C[i].s + \alpha$

\For{$\tau$ in $T$}
\If{$\tau$ is not in $C[i].t$}
\State \hl{$\ell_i(w:{\tt \textless \tau \textgreater}) = -\inf$}
\EndIf

\State
\If{$C[i].t'$ is $[]$ or $\tau \neq C[i].t'[-1]$}
\State \hl{$\ell_i(w:{\tt \textless / \tau \textgreater}) = -\inf$}
\EndIf
\EndFor

\State
\If{$C[i].t$ is not $[]$ or $C[i].t'$ is not $[]$}
\State \hl{$\ell_i(w:{\tt EOS}) = -\inf$}
\EndIf

\EndIf
\EndFor

\State
\State $C' = []$ \Comment{Updated candidates}
\For{$i$ in $1\ldots B$}
\State $w_i, j_i = \underset{w, j}{\arg\max}(\ell_1, \ldots, \ell_j, \ldots, \ell_B)$

\If{$C[j_i].y[-1]$ is ${\tt EOS}$}
\State $C'$.append$(C[j_i])$
\State $\ell_{j_i} = 0$
\State {\bf continue}
\EndIf

\State
\State $y=C[j_i].y+[w_i]$
\State $s=\ell_{j_i}(w:w_i)$
\State $C'$.append$(\{y, s, C[j_i].t, C[j_i].t' \})$

\State
\If{$w_i$ is an XML open tag}
\State $C'[-1].t$.remove$($type of $w_i)$
\State $C'[-1].t'$.append$($type of $w_i)$
\EndIf

\State
\If{$w_i$ is an XML close tag}
\State $C'[-1].t'$.pop$()$
\EndIf

\State
\If{the first token}
\State $\ell_\mathrm{all}(w:w_i) = -\inf$
\Else
\State $\ell_{j_i}(w:w_i) = -\inf$
\EndIf

\EndFor
\State $C = C'$

\EndWhile

\State
\State {\bf return} $C[0].y$

\EndFunction

\end{algorithmic}
}
\end{algorithm}

\section{Detailed Experimental Settings}

This section describes more detailed experimental settings, corresponding to Section~4.

\subsection{Tokenization by Sentencepiece}
We used the SentencePiece toolkit to learn a joint sub-word tokenizer for each language pair, and we set the shared vocabulary size to 8,000 for all the experiments.
In the experiments without the XML tags, the URL placeholders ({\tt \#URL1\#}, {\tt \#URL2\#}, $\ldots$, {\tt \#URL9\#}) are registered as user-defined special symbols when training the tokenizers.
For each of the English-to-\{Japanese, Simplified Chinese\} and Finnish-to-Japanese experiments, we over-sampled English or Finnish text for training the joint sub-word tokenizer, because Japanese and Simplified Chinese have much more unique characters than other alphabetic languages.

In the experiments with XML, we further added all the XML tags (e.g. {\tt \textless b\textgreater}, {\tt \textless /b\textgreater}) to the list of the user-defined special symbols.
We also set the three tokens {\tt \&amp;}, {\tt \&lt;}, and {\tt \&gt;} as the special tokens.
When computing BLEU scores, {\tt \&amp;}, {\tt \&lt;}, and {\tt \&gt;} are replaced with {\tt \&}, {\tt \textless}, and {\tt \textgreater}, respectively.

\subsection{Model Training}

We implemented the transformer model with $K=6$ and $d=256$ as a competitive baseline model.
The number of the multi-head attention layer in the transformer model is 8,  and the dimensionality of its internal hidden states is 1024.
For more details about the multi-head attention layer and the internal hidden states, please refer to \citet{transformer}.

For optimization, we used {\it Adam}~\citep{adam} with a modified weight decay and a cosine learning rate annealing~\citep{adamwr}.
The mini-batch size was set to 128, and the weight decay coefficient was set to $1.0\times 10^{-4}$.
A gradient-norm clipping method was used to stabilize the model training, with the clipping size of 1.0.
The initial learning rate is $5.0\times 10^{-4}$, and it is linearly increased to $1.0\times 10^{-3}$ according to the number of iterations in the first 10 epochs of the model training.
Then, the learning rate and the weight decay coefficient are multiplied by the following annealing factor:
\begin{eqnarray}
\eta_i = 0.5 + 0.5 \cos{\left( \frac{i-10}{50-10} \pi \right)},
\end{eqnarray}
where $\eta_i$ is for the $i$-th ($i \geq 10$) epoch of the model training, and ``50'' is the maximum number of the training epochs.
During the model training, a greedy-generation BLEU score without XML is evaluated at every half epoch by using the development set, and the best-performing checkpoint is used for evaluation.


\end{document}